\pdfoutput=1
\documentclass{article}
\usepackage[preprint,nonatbib]{neurips_2025}
\AtBeginDocument{\newgeometry{textheight=9in,textwidth=6.0in,top=1in,headheight=12pt,headsep=25pt,footskip=30pt}}
\usepackage{times}
\usepackage[numbers,sort&compress]{natbib}
\usepackage[hyphens]{url}

\usepackage{caption}
\usepackage{graphicx}
\usepackage{amsmath,amssymb}
\usepackage{booktabs}
\usepackage{multirow}
\usepackage{xspace}
\usepackage{xcolor}
\usepackage{colortbl}
\usepackage[normalem]{ulem}

\newcommand{\Fig}[1]{Figure~\ref{#1}}
\newcommand{\Tbl}[1]{Table~\ref{#1}}
\newcommand{\Sect}[1]{Section~\ref{#1}}
\newcommand{\Sects}[2]{Sections~\ref{#1} and~\ref{#2}}
\newcommand{\Apx}[1]{Appendix~\ref{#1}}
\newcommand{\Equ}[1]{Eq.~\eqref{#1}}

\newcommand{\method}{Reference Forcing\xspace}
\newcommand{\umethod}{\bname{Reference Forcing}\xspace}
\newcommand{\vres}{value correction\xspace}
\newcommand{\Vres}{Value correction\xspace}

\DeclareMathOperator{\softmax}{softmax}

\definecolor{oursbg}{HTML}{EAF2FC}
\definecolor{basebg}{HTML}{F0F0F0}
\newcommand{\best}[1]{\textbf{#1}}
\newcommand{\second}[1]{\underline{#1}}
\newcommand{\up}{$\uparrow$}

\newcommand{\metric}[1]{\textit{#1}}
\newcommand{\bname}[1]{\uline{#1}}

\makeatletter
\renewcommand{\paragraph}{\@startsection{paragraph}{4}{\z@}{5pt \@plus 1pt \@minus 1pt}{-1em}{\normalfont\normalsize\bfseries}}
\makeatother

\makeatletter
\DeclareRobustCommand\onedot{\futurelet\@let@token\@onedot}
\def\@onedot{\ifx\@let@token.\else.\null\fi\xspace}
\def\eg{\emph{e.g}\onedot}
\makeatother

\definecolor{cvprblue}{rgb}{0.21,0.49,0.74}
\usepackage[pagebackref,breaklinks,colorlinks,allcolors=cvprblue]{hyperref}
\hypersetup{pdftitle={Reference Forcing: Long Video Generation with Region-Guided Drift Correction},
  pdfauthor={Sheng Zhao, Haitian Zheng, Weikai Lin, Junjie Zhao, Yuhao Zhu}}

\title{Reference Forcing: Long Video Generation with Region-Guided Drift Correction}

\author{%
  Sheng Zhao$^{1}$\quad Haitian Zheng$^{2}$\quad Weikai Lin$^{1}$\quad Junjie Zhao$^{1}$\quad Yuhao Zhu$^{1}$\\
  $^{1}$University of Rochester\qquad $^{2}$Adobe
}

\begin{document}
\maketitle
\begin{abstract}
Long video generation should keep the subject and background consistent while supporting rich, natural motion.
However, autoregressive models are often distilled with distribution matching distillation (DMD) over short windows, which does not train them to keep this consistency over long ranges.
As a result, the subject and background can drift over time.
Methods that manage the key-value (KV) cache retain or retrieve history to reduce drift.
Yet this history may itself have drifted.
We propose \method, which uses the initial frames as an appearance reference.
We first build a teacher that uses attention masks to force each region of a new frame to attend to the same region in the initial frames.
The teacher finds these regions with off-the-shelf segmentation models.
We then distill this behavior into a separate attention path of the student, which needs no external models at inference.
We further introduce \vres, which corrects drifted values in the KV cache toward the matching values of drift-free initial frames.
Experiments show that \method achieves state-of-the-art subject and background consistency among long-video methods on different base models, and its advantage grows with video length.
Ablations show both components are needed to gain consistency without losing motion.
More visualizations and code are available at \url{https://zhaosheng-thu.github.io/reference-force/}.
\end{abstract}

\begin{figure}[!htb]
\centering
\includegraphics[width=\linewidth]{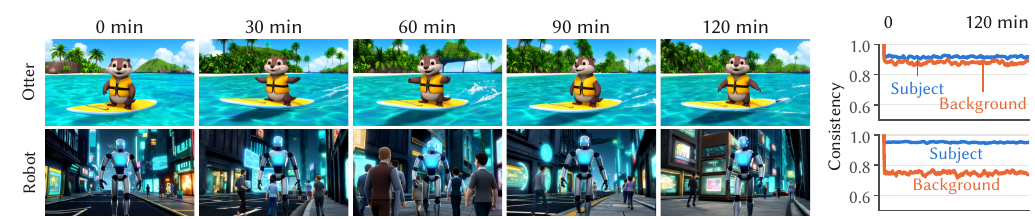}
\caption{%
\textbf{Examples of two-hour videos generated by \method.}
Each row shows frames at 0, 30, 60, 90, and 120 minutes and, on the right, the consistency of the subject and the background, measured by their semantic similarity to the first frame.}
\label{fig:teaser}
\end{figure}

\section{Introduction}
\label{sec:intro}

Recent video diffusion models generate short clips with high visual quality~\cite{svd,cogvideox,hunyuanvideo,wan}.
Autoregressive video diffusion models extend generation to longer videos by generating one chunk at a time, using previously generated frames to guide each new chunk~\cite{diffusionforcing,causvid,selfforcing,magi1,skyreels}.
With few-step distillation and a key-value (KV) cache that stores recent frames, these models can generate videos lasting a minute or longer in real time~\cite{longlive,rollingforcing,selfforcingpp}.
However, long videos must preserve the appearance of subjects and backgrounds while maintaining rich and natural motion.
In practice, errors accumulate during autoregressive generation, gradually changing the appearance of subjects and backgrounds, \eg, a red jacket slowly turns orange~\cite{selfforcing,longlive,memrope}.

One reason for this drift lies in how these models are trained.
LongLive~\cite{longlive}, for example, uses distribution matching distillation (DMD)~\cite{dmd}, which matches the distribution of generated video clips within short windows.
However, this objective does not explicitly require subjects and backgrounds to retain their earlier appearance.
A generated clip can therefore look realistic and receive a low distillation loss even if its appearance has gradually drifted.

Existing methods address this problem by allowing the model to access more of its generation history.
Some preserve the first frames as an attention sink~\cite{streamingllm,longlive}, while others compress older frames~\cite{deepforcing} or retrieve relevant information from previously generated content~\cite{longliverag,memrope}.
However, previously generated frames may already contain appearance errors, so attending to them can propagate drift.
The first frames are free of such errors, but stronger attention to them can freeze the motion~\cite{tethermem}.

To reduce the drift of subjects and backgrounds without sacrificing motion, we propose \method (\Fig{fig:overview}), which combines two complementary designs.

First, we let each region of a new frame attend to the same region of the initial frames.
We build a teacher that finds these regions with off-the-shelf segmentation and matching models and uses attention masks to force each region to attend only to the same region.
We distill this behavior into a separate attention branch of the student, called the anchor path, which attends only to the initial frames and needs no external models at inference.
The student also learns an anchor gate during distillation, which controls how much each token uses the anchor path, depending on whether its content appears in the initial frames.

Second, we correct the drift already stored in the KV cache.
The recent frames may already have drifted, and later frames continue from them.
For each token of a recent frame, our training-free \vres finds the token in the initial frames that shows the same content.
It then moves the cached value of the token toward the value of that token, by the anchor gate learned in distillation.

We evaluate \method on long videos generated from MovieGenBench~\cite{moviegen} prompts and compare it with existing long-video generation methods.
\method achieves the highest subject and background consistency.
It maintains motion dynamics and visual quality comparable to the base model.
\Fig{fig:teaser} shows that it also generates two-hour videos in one pass, with subject and background consistency that stays stable over the two hours.

Our contributions are summarized as follows:
\begin{itemize}
    \item We propose region-guided distillation for long video generation. We build a teacher that uses off-the-shelf models and attention masks to make each region of a new frame attend to the same region of the initial frames, and distill it into an anchor path of the student, which removes these models and their cost at inference.
    \item We introduce \vres, a training-free method for autoregressive video models that corrects the drifted values in the KV cache toward the matching values of the drift-free initial frames.
    \item Extensive experiments show that \method achieves state-of-the-art subject and background consistency, and ablation studies show that each of its components and design choices is needed to improve consistency while keeping the motion of the base model.
\end{itemize}

\section{Related Work}
\label{sec:related}

\paragraph{Autoregressive video generation and distillation.}
Diffusion and flow-based models~\cite{zhao2026gazeflow,ddpm,ldm,dit,attention,flowmatching,rectifiedflow} generate high-quality video clips of several seconds~\cite{svd,cogvideox,hunyuanvideo,opensora,moviegen,wan}, and earlier work extends them to longer videos by hierarchical generation or noise rescheduling~\cite{nuwaxl,freenoise,fifodiffusion,streamingt2v}.
Autoregressive video diffusion generates frames causally and can continue a video beyond its training length~\cite{diffusionforcing,pyramidflow,magi1,skyreels,framepack}.
Building on few-step distillation~\cite{progressivedistill,consistencymodels,dmd,dmd2}, recent models distill a bidirectional teacher into a few-step causal student~\cite{causvid,selfforcing,causalforcing,opsdv} and extend the rollout to minutes~\cite{rollingforcing,selfforcingpp,longlive}, with objectives computed on windows of a few seconds.
Knowledge distillation~\cite{hinton,lin2026lowpowar} also admits teachers with privileged inputs that are unavailable at test time~\cite{lupi}: our teacher uses segmentation~\cite{groundingdino,sam,sam2} and optical flow~\cite{raft}, and the student learns its effect through an additional attention path with low-rank adaptation~\cite{lora}, so it needs neither at test time.

\paragraph{Memory for long video generation.}
Language models keep long contexts in a bounded cache through attention sinks, token eviction, or segment memories~\cite{streamingllm,h2o,transformerxl,memorizing}.
Long video generators adopt similar mechanisms: a frame sink~\cite{longlive}, a deeper and compressed sink~\cite{deepforcing}, positional re-indexing~\cite{infinityrope}, evolving memory tokens~\cite{memrope}, retrieved history~\cite{longliverag,tethermem}, and memory indexed by camera viewpoint~\cite{worldmem,contextasmemory}, all placing the history in the same softmax.
History Forcing~\cite{historyforcing} distills the difference between teacher predictions with and without extra history.
\method attends to the first frames in a separate attention and corrects the values of the recent window.

\section{Method}
\label{sec:method}

\begin{figure*}[t]
\centering
\includegraphics[width=\linewidth]{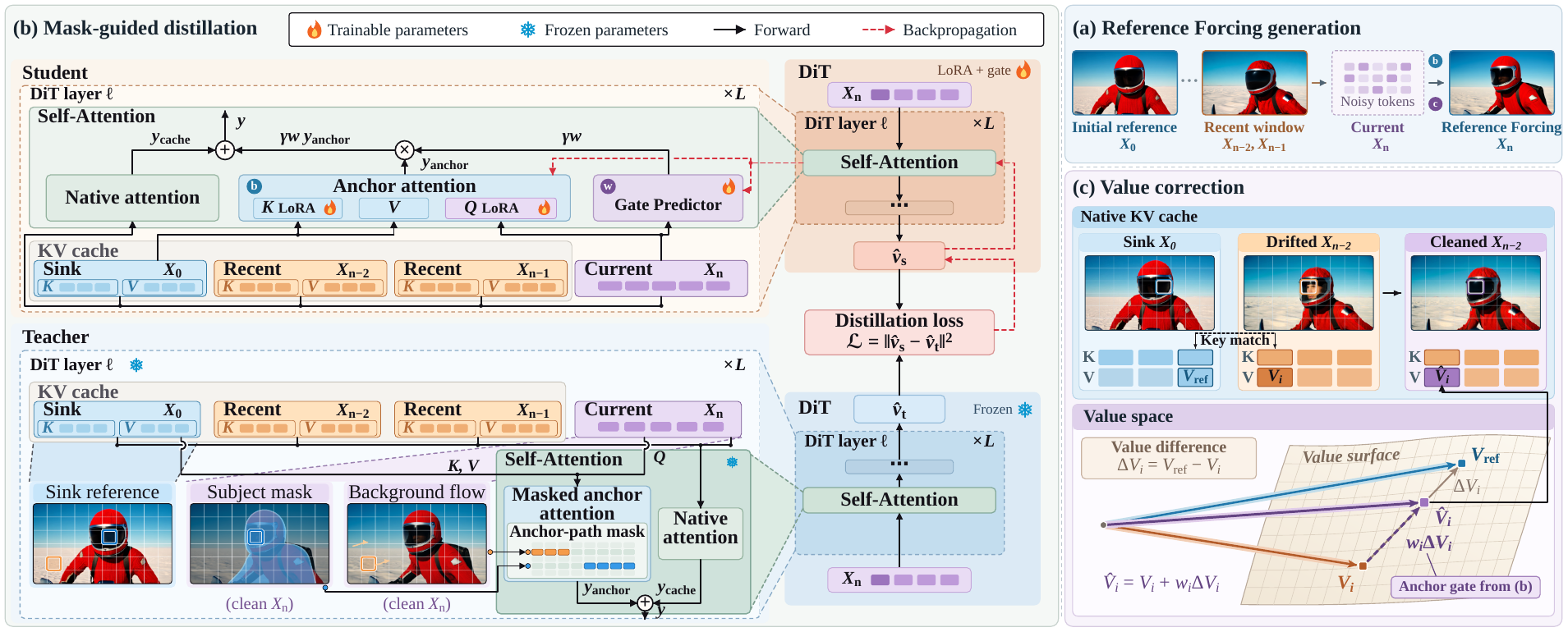}
\caption{%
\textbf{Overview of \method.}
(a) Each chunk $X_n$ is generated from the first chunk $X_0$, the recent window, and noisy tokens, with the anchor path of (b) and the \vres of (c).
(b) In every layer, the student adds an anchor path (LoRA on the query and keys, and the anchor gate $w$) to the frozen native attention and learns to match the prediction of a teacher whose anchor path is masked by the subject mask and the background flow.
(c) \Vres moves the values of the drifted chunk $X_{n-2}$ toward their key-matched values in $X_0$, with the anchor gate $w$ learned in (b) as strength.}
\label{fig:overview}
\vspace{-4pt}
\end{figure*}

\method builds on a frozen autoregressive video generation model, such as LongLive~\cite{longlive} or Self-Forcing~\cite{selfforcing}, and preserves the appearance of the first chunk while new content and motion evolve (\Fig{fig:overview}).
After reviewing the KV cache (\Sect{sec:cache}), we add an anchor path that attends to the first chunk (\Sect{sec:mem}), train it by distilling a teacher with masked attention (\Sect{sec:distill}), and correct the drifted values in the recent cache with \vres (\Sect{sec:vres}).

\subsection{KV Cache of Autoregressive Generation}
\label{sec:cache}

An autoregressive video model generates video chunk by chunk and keeps the keys and values of generated chunks in a KV cache.
In LongLive~\cite{longlive}, for example, each chunk has three latent frames.
As shown in \Fig{fig:overview}(a), when generating the current chunk $X_n$, each self-attention layer attends to the first chunk, two recent chunks, and the current chunk:
\begin{equation}
\mathcal{C}_n = \bigl[\, \underbrace{X_0}_{\text{sink}} \;;\; \underbrace{X_{n-2} \;;\; X_{n-1}}_{\text{recent window}} \;;\; \underbrace{X_n}_{\text{current}} \,\bigr],
\label{eq:cache}
\end{equation}
where each chunk $X_m$ contributes its keys $K_m$ and values $V_m$.
The first chunk $X_0$, called the \emph{sink}, remains available throughout generation, while the recent window advances as new chunks are generated.
The recent chunks provide local context for motion and new content, but may also carry appearance errors accumulated from earlier generation.

\subsection{Anchor Attention}
\label{sec:mem}

The most direct way to improve consistency is to raise the attention to the first chunk, which has not drifted.
In the native attention, however, the first chunk and the recent chunks share one softmax, so this takes weight from the recent chunks, which carry the context for motion, and trades motion for consistency (\Sect{sec:abl_softmax}).
We therefore introduce an independent \emph{anchor path}, which attends to the first chunk without changing the attention to the recent chunks.

We call the original attention over $\mathcal{C}_n$ the \emph{cache path}.
For a query $q$ in the current chunk, each self-attention layer combines the two paths (\Fig{fig:overview}(b)):
\begin{equation}
\begin{aligned}
y_{\text{cache}} &= W_o\, \softmax\!\bigl(q K_{\mathcal{C}}^{\top} / \sqrt{d}\bigr)\, V_{\mathcal{C}}, \\
y_{\text{anchor}} &= W_o\, \softmax\!\bigl(\tilde q\, \tilde K_0^{\top} / \sqrt{d} + M\bigr)\, V_0, \\
y &= y_{\text{cache}} + \gamma\, w\, y_{\text{anchor}},
\end{aligned}
\label{eq:mem}
\end{equation}
where $K_{\mathcal{C}}$ and $V_{\mathcal{C}}$ are the keys and values of $\mathcal{C}_n$, $K_0$ and $V_0$ are those of the first chunk, $W_o$ is the native output projection, and $\gamma$ is a fixed scale.
In the native attention, the rotary positional encoding (RoPE)~\cite{roformer} is applied to the query and keys to encode when and where each token appears.
The anchor path instead takes the query and keys before this step and adds no positional encoding, so a token is matched to the first chunk by content alone, wherever it has moved.

Intuitively, the anchor path lets each token attend to the tokens of the first chunk that show the same content and add their appearance to its output.
In \Fig{fig:overview}(b), for example, the astronaut attends to the astronaut in the first chunk, and a patch of the ground attends to the same patch of the ground.
The query $\tilde q$, the keys $\tilde K_0$, and the mask $M$, which is $0$ where the token may attend and $-\infty$ elsewhere, decide \emph{which} tokens it attends to, and the \emph{anchor gate} $w \in [0,1]$ decides \emph{how much} it adds.
The gate is needed because content that appears later, such as a subject entering the frame, has no matching region in the first chunk.
Attending to the first chunk would copy unrelated appearance onto it, so the gate closes ($w \approx 0$) and leaves it to the native attention.

\subsection{Region-Guided Distillation}
\label{sec:distill}

\paragraph{Teacher.}
\label{sec:teacher}
The teacher uses the mask $M$ to make each token of the current chunk attend only to its matching region in the first chunk.
We segment the subject with Grounding DINO~\cite{groundingdino} and SAM~2~\cite{sam2}, and let a subject token attend only to the same subject in the first chunk.
We match the background by optical flow~\cite{raft}, and let a background token attend only to its matching region in the first chunk.
Both use the anchor path with $w = 1$, and content that does not appear in the first chunk has $w = 0$.
The teacher uses the native query and keys ($\tilde q = q$, $\tilde K_0 = K_0$).
The teacher must decode every chunk and run these off-the-shelf models on it, which makes generation much slower.

\paragraph{Student.}
We therefore distill the teacher into a student that attends to the whole first chunk ($M = 0$) without these models.
We apply low-rank adaptations (LoRA)~\cite{lora} to the query and keys to let each token find its matching region in the first chunk by itself.
The adapted query and keys are $\tilde q$ and $\tilde K_0$, with LoRA weights $\Delta_q$ and $\Delta_k$.
We use the anchor gate to let tokens that do not appear in the first chunk turn the anchor path off.
The gate is predicted from the query by a learned linear layer $(u, b)$ as $w = \sigma\bigl(u^{\top} \mathrm{LN}(q) + b\bigr)$.
\Fig{fig:overview}(b) shows how the teacher (bottom) is distilled into the student (top).

\paragraph{Objective.}
The teacher and the student share the frozen model and differ only in the anchor path, so matching their predictions teaches the anchor path of the student to attend as the teacher does.
The trainable parameters are $\theta = \bigcup_{\ell} \{\Delta_q^{\ell}, \Delta_k^{\ell}, u^{\ell}, b^{\ell}\}$, i.e. the LoRA and the gate in every layer $\ell$.
We train them to minimize the difference between the student and teacher predictions on the same input:
\begin{equation}
\min_{\theta}\; \bigl\lVert \hat{v}_{\text{student}}(x_t, t, \mathcal{C}_n; \theta) - \hat{v}_{\text{teacher}}(x_t, t, \mathcal{C}_n) \bigr\rVert_2^2,
\label{eq:loss}
\end{equation}
where $\hat{v}$ denotes the velocity predicted by the flow-matching model, $X_n$ is a chunk of a generated video with cache $\mathcal{C}_n$, $M$ is computed on $X_n$, and $x_t$ is $X_n$ noised to a level $t$ at which the anchor path is used.

\subsection{Repairing Drifted History}
\label{sec:vres}

The native attention still attends to the recent chunks, whose appearance may already have drifted.
\Vres moves their cached values toward the matching values of the first chunk and keeps their keys (\Fig{fig:overview}(c)).
The native attention thus attends to the same tokens as before but reads a corrected appearance from them.

We match tokens between a recent chunk $X_m$ and the first chunk by mutual nearest neighbors in key similarity.
Two tokens are matched only if each is the most similar token to the other, which keeps only reliable matches and gives a matching $\pi_m$.
We then move the value of each matched token $i$ toward its match:
\begin{equation}
\hat V_m[i] = V_m[i] + w_m[i] \bigl(V_0[\pi_m(i)] - V_m[i]\bigr),
\label{eq:vres}
\end{equation}
where $w_m[i]$ is the anchor gate that the student predicted for token $i$ when $X_m$ was generated, stored in each layer.
Each matched value is thus interpolated toward the value of its match in the first chunk, with the anchor gate as the weight.
The anchor gate is large for the subject and the background that also appear in the first chunk, and small for new content, so new content is barely changed.

We correct only $X_{n-2}$, because the current chunk continues from the last frames of $X_{n-1}$, and correcting them would cause severe flicker artifacts between chunks (\Sect{sec:abl_vres}).
The current chunk therefore attends to
\begin{equation}
\hat{\mathcal{C}}_n = \bigl[\, X_0 \;;\; (K_{n-2}, \hat V_{n-2}) \;;\; (K_{n-1}, V_{n-1}) \;;\; X_n \,\bigr].
\label{eq:vcache}
\end{equation}

\section{Experimental Setup}
\label{sec:exp}
\label{sec:setup}

\begin{table*}[t]
\centering
\caption{%
\textbf{Comparison on MovieGenBench.}
VBench-Long scores of 30-, 60-, and 120-second videos; all methods use the same KV cache capacity (12 latent frames) and start from the same noise.
Rank: average rank over the six columns.
\textbf{Bold}: best; \underline{underline}: second.}
\label{tab:main}
\footnotesize
\renewcommand{\arraystretch}{1.0}
\setlength{\tabcolsep}{3.3pt}
\begin{tabular}{c l cc cc cc c}
\toprule
\multirow{2}{*}{Length} & \multirow{2}{*}{Method} & \multicolumn{2}{c}{Consistency} & \multicolumn{2}{c}{Motion} & \multicolumn{2}{c}{Frame quality} & Overall \\
\cmidrule(lr){3-4}\cmidrule(lr){5-6}\cmidrule(lr){7-8}\cmidrule(lr){9-9}
 & & Subject\up & Background\up & Smoothness\up & Dynamic\up & Aesthetic\up & Imaging\up & Rank \\
\midrule
\rowcolor{basebg}
\cellcolor{white}\multirow{6}{*}{30\,s} & LongLive~\cite{longlive}           & \second{97.47} & \second{96.56} & 98.68 & 42.81 & \second{60.98} & 68.67 & \second{2} \\
 & + MemRoPE~\cite{memrope}           & 97.34 & 96.54 & \second{98.70} & 44.69 & 60.90 & \second{69.25} & \second{2} \\
 & + LongLive-RAG~\cite{longliverag}  & 97.27 & 96.40 & 98.65 & \best{48.59} & 60.78 & \best{69.27} & 4 \\
 & + $\infty$-RoPE~\cite{infinityrope} & 97.11 & 96.42 & 98.68 & \second{47.97} & 60.44 & 68.55 & 5 \\
 & + Deep Forcing~\cite{deepforcing}  & 97.13 & 96.33 & 98.35 & 41.88 & 60.92 & 68.89 & 6 \\
\rowcolor{oursbg}
\cellcolor{white} & + \method                          & \best{97.55} & \best{96.70} & \best{98.73} & 41.88 & \best{61.03} & 68.17 & \best{1} \\
\midrule
\rowcolor{basebg}
\cellcolor{white}\multirow{6}{*}{60\,s} & LongLive~\cite{longlive}           & \second{97.33} & 96.49 & \second{98.67} & 42.19 & 60.76 & 68.58 & 3 \\
 & + MemRoPE~\cite{memrope}           & 97.32 & \second{96.50} & 98.59 & 46.80 & 60.15 & \best{69.08} & \second{2} \\
 & + LongLive-RAG~\cite{longliverag}  & 97.31 & 96.38 & 98.61 & \second{49.53} & 60.15 & \second{68.88} & 4 \\
 & + $\infty$-RoPE~\cite{infinityrope} & 97.03 & 96.36 & 98.54 & \best{51.09} & 59.35 & 68.29 & 6 \\
 & + Deep Forcing~\cite{deepforcing}  & 97.31 & 96.48 & 98.24 & 42.89 & \second{60.78} & 68.79 & 5 \\
\rowcolor{oursbg}
\cellcolor{white} & + \method                          & \best{97.56} & \best{96.67} & \best{98.68} & 43.13 & \best{60.89} & 68.10 & \best{1} \\
\midrule
\rowcolor{basebg}
\cellcolor{white}\multirow{6}{*}{120\,s} & LongLive~\cite{longlive}           & 97.06 & 96.30 & \second{98.59} & 46.45 & \best{59.84} & 68.27 & \best{1} \\
 & + MemRoPE~\cite{memrope}           & 96.92 & \second{96.34} & 98.54 & 48.12 & 59.57 & \best{68.91} & 4 \\
 & + LongLive-RAG~\cite{longliverag}  & \second{97.09} & 96.25 & 98.57 & 49.92 & 59.57 & \second{68.30} & \second{3} \\
 & + $\infty$-RoPE~\cite{infinityrope} & 96.72 & 96.20 & 98.52 & \best{52.73} & 58.91 & 68.02 & 6 \\
 & + Deep Forcing~\cite{deepforcing}  & 96.93 & 96.18 & 98.49 & \second{50.94} & 59.35 & 68.21 & 5 \\
\rowcolor{oursbg}
\cellcolor{white} & + \method                          & \best{97.27} & \best{96.56} & \best{98.61} & 44.69 & \second{59.83} & 67.98 & \best{1} \\
\midrule
\rowcolor{basebg}
\cellcolor{white}\multirow{6}{*}{30\,s} & Self-Forcing~\cite{selfforcing}    & 96.57 & 95.89 & 97.96 & 40.86 & 58.56 & 68.58 & 5 \\
 & + MemRoPE~\cite{memrope}           & 96.94 & 96.23 & 97.91 & 61.95 & 58.86 & \second{68.87} & \second{3} \\
 & + LongLive-RAG~\cite{longliverag}  & \best{97.13} & \second{96.28} & \second{98.01} & 60.16 & \best{59.62} & \best{69.11} & \best{1} \\
 & + $\infty$-RoPE~\cite{infinityrope} & 96.84 & 96.20 & 97.94 & \second{64.61} & 58.52 & 68.75 & 4 \\
 & + Deep Forcing~\cite{deepforcing}  & 96.96 & 96.15 & 97.86 & 54.77 & 57.29 & 68.04 & 6 \\
\rowcolor{oursbg}
\cellcolor{white} & + \method                          & \second{96.99} & \best{96.30} & \best{98.04} & \best{65.08} & \second{59.06} & 68.65 & \best{1} \\
\bottomrule
\end{tabular}
\end{table*}

\paragraph{Training and inference.}
We build on \bname{LongLive}-1.3B~\cite{longlive} and keep all of its weights frozen.
Only the anchor path is trained (\Sect{sec:distill}), for 20{,}000 steps.
The training data are 60-second videos generated mainly by \bname{LongLive} for 2{,}000 prompts from MovieGenBench and VidProM~\cite{vidprom}, which exclude the test prompts and prompts close to them in meaning.
Each training sample is one chunk of such a video with the cache used to generate it, and the teacher masks are computed on the clean chunk before it is noised.
We additionally test \bname{Self-Forcing} 1.3B~\cite{selfforcing}, on which the anchor path is distilled again.\footnote{\bname{Self-Forcing} is distilled on 5-second clips and cannot generate long videos on its own, so we adapt \umethod to it by adding a sink.}
At inference, the anchor path is used at the last three of the four denoising steps of each chunk.
\Apx{sec:supp_method} gives further details.

\paragraph{Benchmark and baselines.}
We generate 60-second videos ($832\times480$, 16 frames per second) for the 128 MovieGenBench prompts~\cite{moviegen, memrope}, starting every method from the same initial noise.
For efficiency, the ablations and analyses in \Sects{sec:abl}{sec:analysis} use the 60-second videos of a randomly sampled half of these prompts.
We compare with \bname{LongLive}~\cite{longlive} and four long-video methods applied to it: \bname{MemRoPE}~\cite{memrope}, \bname{LongLive-RAG}~\cite{longliverag}, \bname{$\infty$-RoPE}~\cite{infinityrope}, and \bname{Deep Forcing}~\cite{deepforcing}.
Each baseline is run with its released code.
On \bname{Self-Forcing}, all methods generate 30-second videos for the same prompts.

\paragraph{Overall metrics.}
To evaluate overall consistency, motion, and frame quality, we use six dimensions of VBench-Long~\cite{vbench,vbenchpp} and their average rank~\cite{longliverag}:
\metric{Subject consistency} and \metric{background consistency} measure whether the appearance of the subject and of the background stays the same over the video, by comparing the features of each frame with those of the first and the previous frame.
\metric{Motion smoothness} measures whether the motion is smooth and free of jitter, and \metric{dynamic degree} measures how much the video moves.
\metric{Aesthetic quality} and \metric{imaging quality} measure the visual appeal of each frame and its distortions, such as noise and blur, with the aesthetic and image-quality predictors of VBench-Long~\cite{vbench,vbenchpp}.

\paragraph{Long-horizon evaluation.}
To measure long-horizon consistency, we compute the DINOv2~\cite{dinov2} similarity of the subject in each frame to the subject in the first frame (following VBench), and likewise for the background, and average the subject similarity over the last 20 seconds of the 60-second videos, which we call \metric{L-Consistency} (L-Cons.).
To see how consistency evolves, we further plot both similarities over 120-second videos of the same prompts (like \Fig{fig:curves}).
We measure motion by the \metric{tracking strength} of MiraBench~\cite{miradata}.\footnote{Unlike the dynamic degree of VBench-Long, which is computed from optical flow, tracking strength does not mistake the appearance change of a drifting subject for motion~\cite{snfbench}.}

\paragraph{Pairwise evaluation.}
To compare the consistency, motion, and quality of the generated videos as people perceive them, we conduct a user study~\cite{zhao2026observer,lin2026lowpowar} and a pairwise evaluation with a vision-language model (VLM, Claude Opus 5.5) under the same two-alternative forced-choice (2AFC) protocol, following \bname{Deep Forcing}~\cite{deepforcing}.
We recruit 7 participants under IRB approval, each of whom watches the 60-second videos of \umethod and of one other method for the same prompt, shown side by side in random order, and chooses the better one overall and in consistency, motion, and quality.
The VLM judge answers the same questions.

\section{Results}
\label{sec:ablation}

\Sect{sec:results} compares \umethod with prior methods.
\Sect{sec:abl} ablates its components and design choices, and \Sect{sec:analysis} examines what the anchor path learns.

\subsection{Main Results}
\label{sec:results}

\begin{figure*}[t]
\centering
\includegraphics[width=\linewidth]{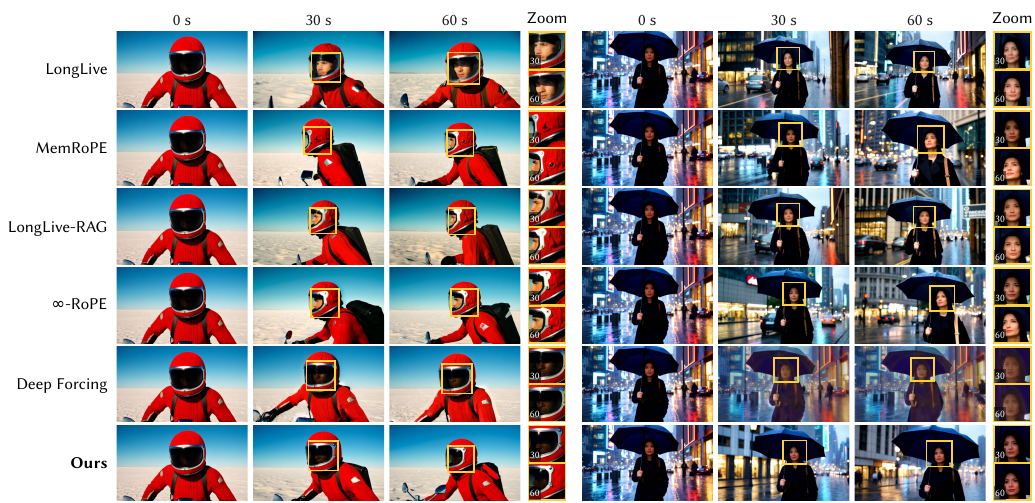}
\caption{%
\textbf{Qualitative comparison.}
All methods share the prompt and first chunk; boxes mark subject changes, magnified in the zoom column.
Left: the dark visor of the astronaut turns transparent in baselines but stays dark in \method.
Right: in baselines, the face of the woman becomes lighter and the strap on her left shoulder wider and brighter, whereas \method keeps both.
}
\label{fig:qual}
\end{figure*}

\begin{table*}[t]
\centering
\caption{%
\textbf{Pairwise evaluation by VLM judge and by user study.}
Percentage of comparisons in which \method is preferred over each method on MovieGen prompts, with 95\% confidence intervals. Bold marks values whose interval lies above 50.
Both prefer \method in consistency and motion against most methods.
}
\label{tab:vlm}
\footnotesize
\renewcommand{\arraystretch}{1.1}
\setlength{\tabcolsep}{3.6pt}
\begin{tabular}{l cccc cccc}
\toprule
 & \multicolumn{4}{c}{VLM judge} & \multicolumn{4}{c}{User study} \\
\cmidrule(lr){2-5}\cmidrule(lr){6-9}
\method vs. & Overall & Consistency & Motion & Quality & Overall & Consistency & Motion & Quality \\
\midrule
LongLive & \best{78} {\scriptsize [67,\,86]} & \best{86} {\scriptsize [77,\,93]} & \best{78} {\scriptsize [67,\,86]} & \best{63} {\scriptsize [51,\,73]} & \best{72} {\scriptsize [64,\,80]} & \best{76} {\scriptsize [68,\,83]} & \best{72} {\scriptsize [64,\,80]} & \best{73} {\scriptsize [64,\,81]} \\
MemRoPE~\cite{memrope} & \best{72} {\scriptsize [61,\,82]} & \best{76} {\scriptsize [65,\,85]} & \best{69} {\scriptsize [57,\,79]} & 60 {\scriptsize [48,\,71]} & \best{75} {\scriptsize [67,\,83]} & \best{79} {\scriptsize [70,\,85]} & \best{73} {\scriptsize [64,\,81]} & \best{72} {\scriptsize [64,\,80]} \\
LongLive-RAG~\cite{longliverag} & 58 {\scriptsize [46,\,68]} & \best{69} {\scriptsize [57,\,79]} & \best{64} {\scriptsize [52,\,74]} & 44 {\scriptsize [33,\,55]} & \best{74} {\scriptsize [65,\,81]} & \best{79} {\scriptsize [70,\,85]} & \best{71} {\scriptsize [63,\,79]} & \best{75} {\scriptsize [66,\,82]} \\
$\infty$-RoPE~\cite{infinityrope} & \best{84} {\scriptsize [74,\,91]} & \best{85} {\scriptsize [75,\,92]} & \best{73} {\scriptsize [61,\,82]} & \best{73} {\scriptsize [61,\,82]} & \best{79} {\scriptsize [71,\,86]} & \best{83} {\scriptsize [75,\,89]} & \best{75} {\scriptsize [66,\,82]} & \best{76} {\scriptsize [68,\,83]} \\
Deep Forcing~\cite{deepforcing} & \best{78} {\scriptsize [67,\,86]} & 39 {\scriptsize [28,\,50]} & \best{99} {\scriptsize [93,\,100]} & \best{86} {\scriptsize [77,\,93]} & \best{90} {\scriptsize [84,\,95]} & \best{87} {\scriptsize [79,\,92]} & \best{90} {\scriptsize [83,\,94]} & \best{90} {\scriptsize [83,\,94]} \\
\midrule
All & \best{74} {\scriptsize [69,\,78]} & \best{71} {\scriptsize [66,\,75]} & \best{76} {\scriptsize [72,\,80]} & \best{65} {\scriptsize [60,\,70]} & \best{78} {\scriptsize [75,\,81]} & \best{80} {\scriptsize [77,\,84]} & \best{76} {\scriptsize [73,\,79]} & \best{77} {\scriptsize [74,\,80]} \\
\bottomrule
\end{tabular}
\end{table*}

\paragraph{Quantitative and qualitative results.}
\umethod ranks first overall on VBench-Long.
As shown in \Tbl{tab:main}, at both 30 and 60 seconds it achieves the highest subject consistency, background consistency, motion smoothness, and aesthetic quality, and thus the best average rank, while its dynamic degree stays on par with that of \bname{LongLive}.
On \bname{Self-Forcing}, \umethod has the highest background consistency, motion smoothness, and dynamic degree.
Qualitatively, \umethod better preserves the fine appearance of the subject, such as its colors and parts, while the subject moves.
For example, in \Fig{fig:qual}, \bname{LongLive}, \bname{MemRoPE}, \bname{LongLive-RAG}, and \bname{$\infty$-RoPE} turn the dark visor of the astronaut transparent and lighten the woman's face, and \bname{Deep Forcing} turns the whole frame purple and blurry.
\umethod keeps the dark visor with its silver rim and the face's skin tone and shading as they appear in the first frame.
More comparisons are in \Apx{sec:supp_vis} and on the project page.

\begin{figure*}[t]
\centering
\includegraphics[width=0.93\linewidth]{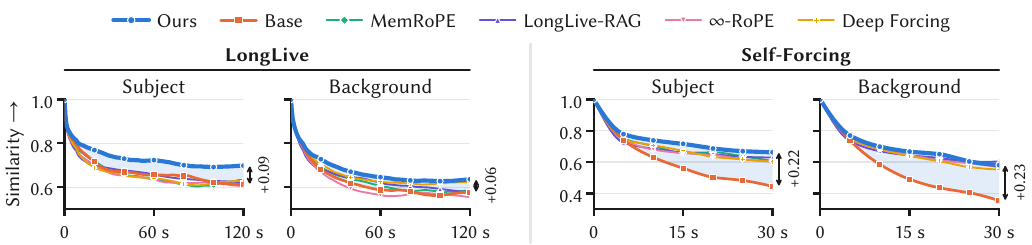}
\caption{%
\textbf{Similarity to the first frame.}
DINOv2 similarity of the subject and the background to the first frame, averaged over MovieGen prompts, on LongLive (120 seconds) and on Self-Forcing (30 seconds).
Base: LongLive and Self-Forcing without a long-video method.
Arrows: the gap between \method and Base at the last time.}
\label{fig:curves}
\vspace{-8pt}%
\end{figure*}

\paragraph{Long-horizon consistency.}
\umethod stays closer to the first frame than \bname{LongLive}, its base model, throughout the two minutes, and the gap widens with time.
In \Fig{fig:curves}, all methods lose a similar amount of consistency in the first five seconds, as the subject starts to move.
Afterwards, \umethod loses about 30\% less than \bname{LongLive} for the subject and 25\% less for the background, and over the last 20 seconds its subject stays closer to the first frame than in all other methods.

\paragraph{User study and VLM evaluation.}
As shown in \Tbl{tab:vlm}, the participants prefer \umethod over every compared method in consistency, motion, and quality, and the VLM judge shows a similar overall preference.

\begin{figure*}[t]
\centering
\begin{minipage}[t]{0.508\linewidth}
\centering
\includegraphics[width=\linewidth]{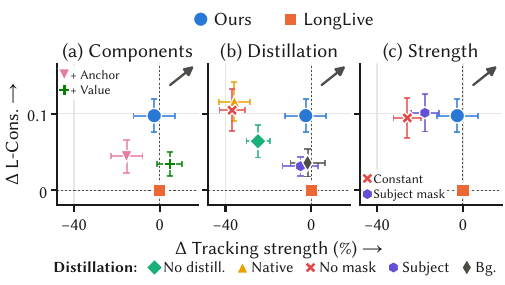}
\caption{%
\textbf{Consistency against motion.}
Change in L-Consistency and tracking strength relative to LongLive (origin), with standard error.
(a) The anchor path and \vres added one at a time.
(b) Other ways to train the anchor path.
(c) Other strengths of \vres.
Only \method raises L-Consistency substantially while keeping the motion of LongLive.}
\label{fig:tradeoff}
\end{minipage}\hfill
\begin{minipage}[t]{0.442\linewidth}
\centering
\includegraphics[width=\linewidth]{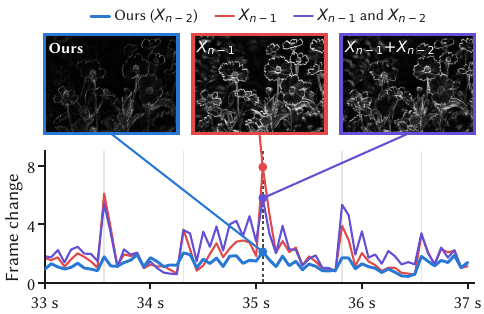}
\caption{%
\textbf{Flicker artifacts from correcting $X_{n-1}$.}
Frame-to-frame change of one video when \vres corrects $X_{n-1}$, or both $X_{n-1}$ and $X_{n-2}$, instead of $X_{n-2}$ only; gray lines mark chunk boundaries, and the images show the difference between the two frames at the dashed one.
Correcting $X_{n-1}$ causes flicker at the chunk boundaries.}
\label{fig:flicker}
\end{minipage}
\end{figure*}

\begin{figure*}[t]
\centering
\includegraphics[width=\linewidth]{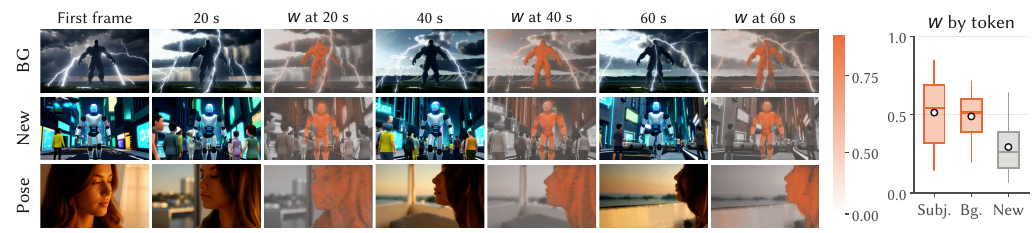}
\caption{%
\textbf{Anchor gate $w$ on generated videos.}
Left: frames and $w$ (orange) at 20, 40, and 60 seconds for a changing background (BG), appearing new subjects (New), and a changing pose (Pose).
Right: distribution of $w$ for subject tokens (Subj.), background tokens also visible in the first chunk (Bg.), and new content (New).
$w$ stays high on the subject and low on new content.}
\label{fig:gate}
\end{figure*}

\begin{figure*}[t]
\centering
\includegraphics[width=\linewidth,trim=0 3pt 0 0,clip]{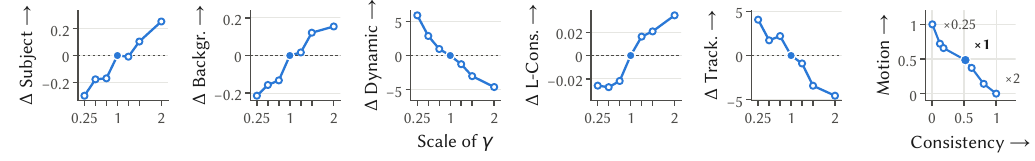}\\[-0.25em]
\includegraphics[width=\linewidth]{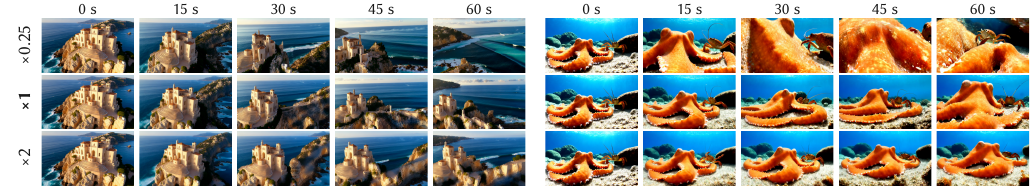}
\caption{%
\textbf{Gain of the anchor path.}
Top left: change of each metric relative to the default gain of 1.
Top right: consistency (mean of subject, background, and L-Consistency) against motion (mean of dynamic degree and tracking strength), each normalized over the seven gains; the upper right is better.
Bottom: the same prompt and initial noise with different gains.
A higher gain gives more consistency and less motion: a low gain loses the church and the octopus, a high gain nearly freezes the scene, and the default keeps both moving.}
\label{fig:sweep}
\end{figure*}

\subsection{Ablation Study}
\label{sec:abl}

We measure each variant by how much it changes L-Consistency and tracking strength relative to \bname{LongLive}.
Both metrics are needed, since a video can also stay close to its first frame by moving less.
As shown in \Fig{fig:tradeoff}, \umethod is the only variant that raises consistency substantially while keeping the motion of \bname{LongLive}.

\paragraph{Components.}
\label{sec:abl_both}
To measure what each component contributes, we add the anchor path and \vres to \bname{LongLive} one at a time.
As shown in \Fig{fig:tradeoff}(a), the two components are complementary.
Added to \bname{LongLive} alone, the anchor path raises consistency but reduces motion ($-15.5\%$), and \vres slightly increases motion but raises consistency less.
Together, they double the gain of either component (L-Consistency $+0.097$ against $+0.044$ and $+0.034$) while keeping the motion of \bname{LongLive}.
The two act on different sources of drift.
The anchor path brings the appearance of the first chunk into the current chunk, while \vres removes the drift in the recent cache.

\paragraph{Distillation.}
\label{sec:abl_softmax}
\label{sec:abl_mask}
To find how the anchor path should be trained, we compare it with variants that skip distillation, distill into the native attention, or use a teacher with only part of its mask or none.
As \Fig{fig:tradeoff}(b) shows, the anchor path keeps the motion only when it is distilled from the masked teacher into its own softmax.
Without distillation, the anchor path reduces motion and raises consistency less (tracking strength $-24.8\%$, L-Consistency $+0.064$).
Distilling the same teacher into the native attention gives the highest consistency but reduces motion significantly ($-35.9\%$), because the first chunk and the recent window then share one softmax, and any attention given to the first chunk is taken from the recent window, which carries the motion.
A teacher without a mask reduces motion as much ($-36.9\%$), because new content then also attends to the first chunk and is pulled toward its appearance.
Keeping only the subject mask or only the background flow preserves the motion but raises consistency much less ($+0.031$ and $+0.036$ against $+0.097$), so both parts of the mask are needed.

\paragraph{Value correction design.}
\label{sec:abl_vres}
To test how \vres should be designed, we vary its strength, whether it also changes the keys, and which recent chunk it corrects.
As \Fig{fig:tradeoff}(c) shows, replacing $w$ with a constant strength or a subject mask from segmentation gives about the same consistency but reduces motion ($-25.9\%$ and $-17.7\%$), because both also correct content that should move.
Correcting the keys as well lowers consistency without changing motion, so changing only the values is enough.
Correcting the newest chunk $X_{n-1}$, alone or with $X_{n-2}$, raises consistency further but causes flicker (\Fig{fig:flicker}).

Overall, three choices keep the motion of \umethod: the anchor path has its own softmax, \vres leaves the newest chunk untouched, and its strength follows $w$.
Replacing any of them reduces motion by 16--36\% relative to \bname{LongLive}.

\subsection{Analysis}
\label{sec:analysis}

\paragraph{Anchor gate.}
\label{sec:abl_learn}

The anchor gate $w$ decides which tokens use the anchor path.
To check whether $w$ opens for the right tokens, we measure it on held-out teacher videos, where the masks of the teacher mark which tokens should use the anchor path.
As shown in the right panel of \Fig{fig:gate}, $w$ is high on the subject and low on content that does not appear in the first chunk.
On videos generated by \umethod (left panel), $w$ follows the subject as it moves and stays low on content that appears later.
The anchor path is thus used mostly by content that also appears in the first chunk, which is why $w$ also sets the strength of \vres (\Sect{sec:vres}).

\paragraph{Anchor path gain.}
\label{sec:abl_gain}
To test how sensitive \umethod is to the strength of the anchor path, we multiply its fixed scale $\gamma$ (\Equ{eq:mem}, \Sect{sec:mem}) by factors from 0.25 to 2, with 1 as the default.
As shown in the top of \Fig{fig:sweep}, a larger $\gamma$ overall raises the subject, background, and long-range consistency and lowers both motion metrics.
The default lies in the middle of the resulting trade-off between consistency and motion (top right).
In the examples (bottom), a small $\gamma$ leads to severe drift, while a large $\gamma$ nearly freezes the scene.

\section{Conclusion and Limitations}
\label{sec:conclusion}

We presented \method, which keeps long videos consistent with their initial frames while preserving motion.
A teacher uses off-the-shelf models and attention masks to make each region of a new frame attend to the same region of the initial frames, and we distill it into an anchor path with a learned anchor gate, which needs no external models at inference.
\Vres further moves the drifted values in the KV cache toward the matching values of the initial frames.
\method achieves the highest subject and background consistency on LongLive while keeping its motion, and its advantage grows with video length.

\paragraph{Limitations.}
\method uses only the first chunk as reference, so content that first appears later, such as an object that enters the scene, has no reference, and updating the reference during generation is left for future work.
Imaging quality remains slightly below that of LongLive, as the anchor path averages over the matched region of the first chunk and can smooth fine textures slightly.
Scenes with several similar subjects remain difficult, because the segmentation used by the teacher follows a single subject.
The teacher also relies on a detector to find the subject in the first chunk.
If the detector finds no subject, the teacher treats the whole frame as background and matches it by optical flow.
This mostly happens in scenes without a main subject, such as landscapes or crowds.

\par

{
    \small
    \bibliographystyle{ieeenat_fullname}
    \bibliography{main,arxiv_restore}
}

\clearpage
\appendix

\section{Method Details}
\label{sec:supp_method}

\subsection{Training}
\label{sec:supp_data}
\paragraph{Prompts.}
We collect 2{,}000 training prompts, 790 from MovieGenBench and 1{,}210 from VidProM~\cite{vidprom}.
We exclude the 128 test prompts and the prompts whose CLIP text embedding has a cosine similarity of at least 0.8 to one of them.
We also exclude VidProM prompts that require rendered text or logos or that name real people or brands.
All training prompts are rewritten with Qwen2.5-7B-Instruct~\cite{qwen25} and the prompt-extension template of Wan2.1~\cite{wan}.

\paragraph{Model.}
The anchor path uses $\gamma = 0.1$ and LoRA of rank 16 on the query and the keys.
The native-attention student of \Sect{sec:abl_softmax} trains a rank-16 LoRA on the query and key projections of the native attention, at the same three denoising steps and with the same data and schedule.

\paragraph{Optimization.}
In one training step, we noise $X_n$ to $x_t$, the teacher predicts the velocity from $(x_t, t, \mathcal{C}_n)$ with its mask $M$, the student predicts it from the same input with $M = 0$, and we update $\theta$ with \Equ{eq:loss}.
Each sample is noised to the three levels at which the anchor path is used ($t = 750$, 500, and 250), with one optimizer step per level.
We use AdamW with a learning rate of $10^{-4}$, no weight decay, and gradient clipping at 1, and train for 20{,}000 steps with one sample per step.

\subsection{Teacher}
\label{sec:supp_teacher}
\paragraph{Subject.}
We detect the subject named in the prompt with Grounding DINO~\cite{groundingdino} and segment it with SAM~2~\cite{sam2}, on the first frame of the first chunk and on one frame of each training chunk.
A token belongs to the subject if at least half of its pixels lie in the mask.
A subject token of the current chunk attends only to the subject tokens of the first chunk, among which the attention selects the matching part by content.
If the first chunk has no subject, the whole frame is treated as background.

\paragraph{Background.}
We compute optical flow with RAFT~\cite{raft} in both directions between a frame of the current chunk and the middle frame of the first chunk, and keep the pixels that pass a forward-backward consistency check.
A background token whose pixels are mostly kept attends to the background tokens of the first chunk around the point that its center maps to.
Background tokens without such a match, and objects that enter later, are not visible in the first chunk and do not use the anchor path.

\subsection{Value Correction}
\label{sec:supp_vres}
\paragraph{Matching.}
\Vres matches tokens by the keys that the model itself computes, so it needs no external model at inference.
For each token, we take its keys before RoPE in the middle blocks of the model (blocks 13 to 18 of 30), normalize them in each block, and concatenate them into one feature.
A token of the recent chunk and a token of the first chunk are matched if each is the most similar token to the other under the cosine similarity of these features.
Tokens without such a match keep their values.
\begin{table}[t]
\centering
\caption{%
\textbf{Matching accuracy of \vres.}
Matches between recent chunks and the first chunk in 64 videos generated by \method, compared with optical flow from RAFT~\cite{raft} with a forward-backward check.
Precision: share of matches within one token of the token that the flow points to.
Coverage: share of the tokens visible in the first chunk under the flow that are matched.}
\label{tab:supp_match}
\footnotesize
\setlength{\tabcolsep}{5pt}
\begin{tabular}{l ccc}
\toprule
Features & Precision (\%) & Coverage (\%) & External model \\
\midrule
Random            & 0.6  & --   & no \\
DINOv2~\cite{dinov2} & 86.5 & 82.5 & yes \\
Model keys (ours) & 86.2 & 78.9 & no \\
\bottomrule
\end{tabular}
\end{table}

To check the matches, we compare them with the correspondences of optical flow on 64 generated videos (\Tbl{tab:supp_match}).
The keys of the model match tokens as accurately as DINOv2 features, without an external model.

\section{More Qualitative Results}
\label{sec:supp_vis}
\begin{figure*}[p]
\centering
\includegraphics[scale=1.5,trim=0 0 225 0,clip]{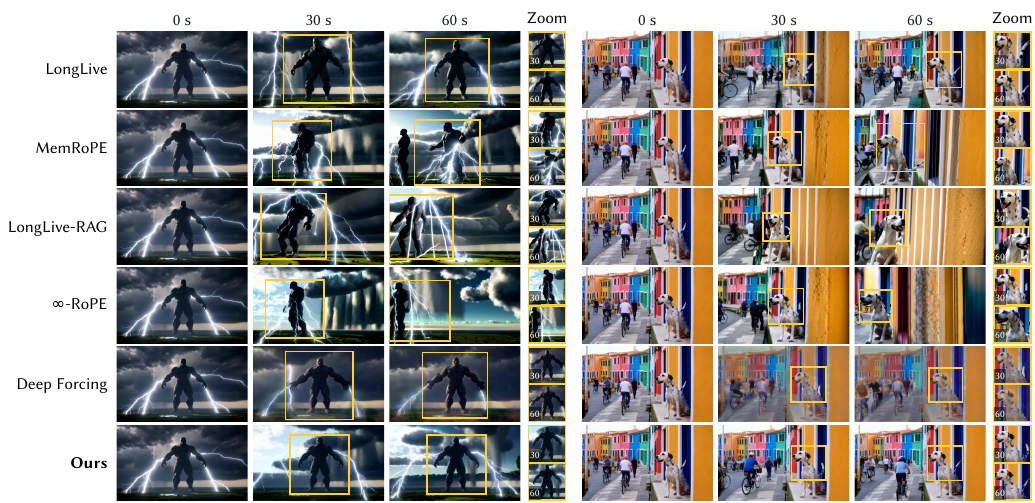}
\caption{%
\textbf{More qualitative comparisons (1/3).}
Layout as in \Fig{fig:qual}.
The massive cloud figure turns into a thin figure in MemRoPE, LongLive-RAG, and $\infty$-RoPE and is blurred in Deep Forcing.
\method keeps its shape, color, and details from the first frame while it still moves.}
\label{fig:supp_qual}
\end{figure*}

\begin{figure*}[p]
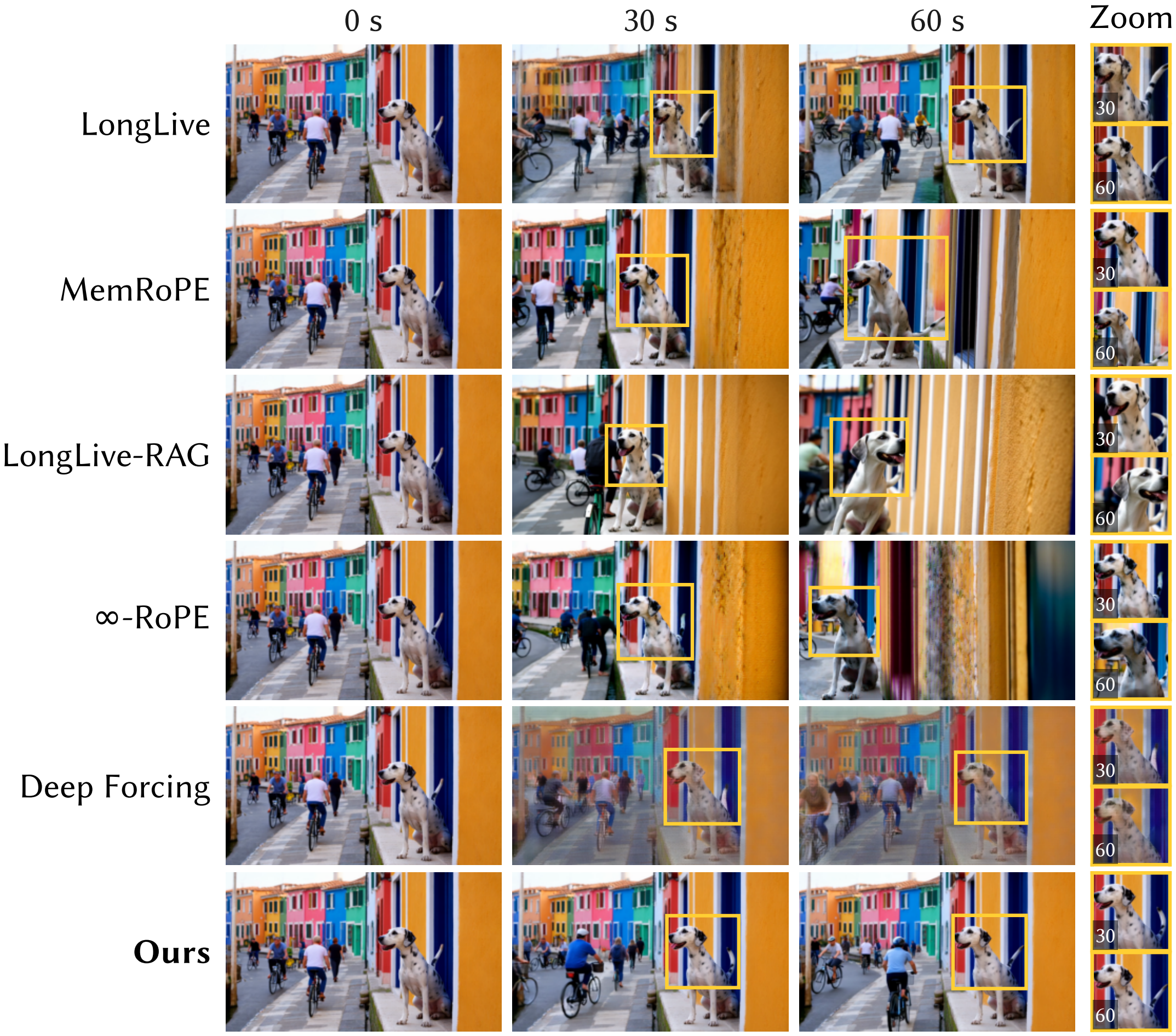

\centering
\includegraphics[scale=1.5,trim=0 0 442 0,clip]{fig/supp_qual_img.pdf}%
\includegraphics[scale=1.5,trim=279 0 0 0,clip]{fig/supp_qual_img.pdf}
\caption{%
\textbf{More qualitative comparisons (2/3).}
Layout as in \Fig{fig:qual}.
The colorful street is replaced by a yellow wall in MemRoPE, LongLive-RAG, and $\infty$-RoPE.
\method keeps the street of the first frame while the camera still moves.}
\label{fig:supp_qual_b}
\end{figure*}

\begin{figure*}[p]
\centering
\includegraphics[scale=1.5,trim=0 0 442 0,clip]{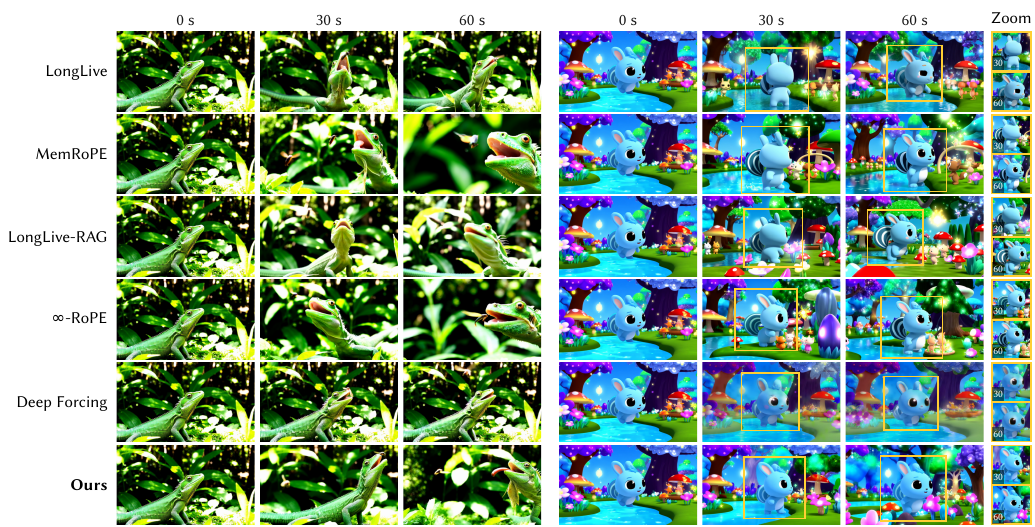}%
\includegraphics[scale=1.5,trim=268 0 0 0,clip]{fig/supp_qual2.pdf}
\caption{%
\textbf{More qualitative comparisons (3/3).}
Layout as in \Fig{fig:qual}.
The small tail of the creature grows several times larger with dark stripes in MemRoPE, LongLive-RAG, and $\infty$-RoPE, and a white belly appears.
\method keeps the shape, color, and details of the first frame while the creature still moves.}
\label{fig:supp_qual_c}
\end{figure*}

\begin{figure*}[t]
\centering
\includegraphics[width=\linewidth,trim=0 216 0 0,clip]{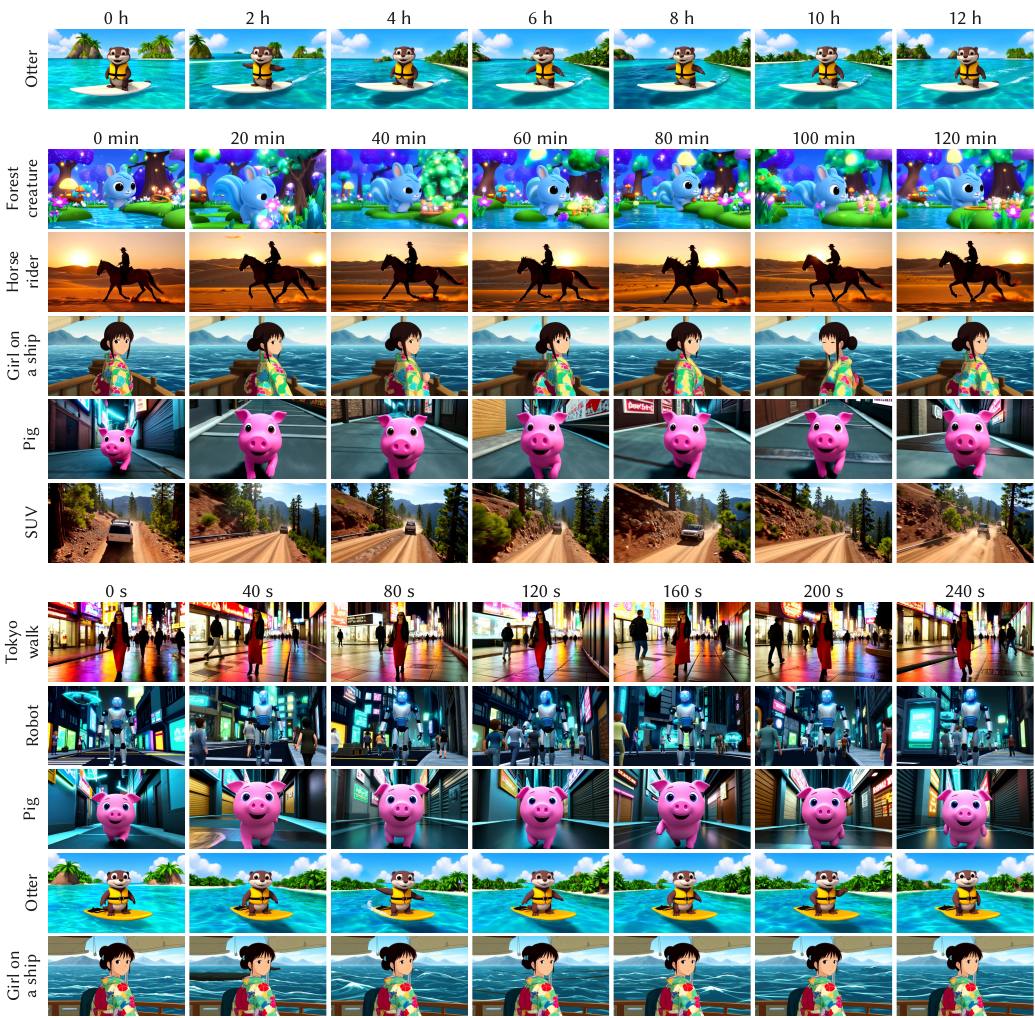}
\caption{%
\textbf{Videos of different lengths generated by \method.}
Each row shows seven frames evenly spaced over the video.
From top to bottom: a 12-hour video and five 2-hour videos, each generated in one pass without restarts or stitching.}
\label{fig:supp_lengths}
\end{figure*}

Figures~\ref{fig:supp_qual} to~\ref{fig:supp_qual_c} compare all methods on three more prompts in the layout of \Fig{fig:qual}.
\Fig{fig:supp_lengths} shows videos of two and twelve hours generated by \method.

\end{document}